\documentclass[runningheads]{llncs}

\usepackage{eccv}

\usepackage{eccvabbrv}

\usepackage{graphicx}
\usepackage{booktabs}

\usepackage{mathtools}
\usepackage{multirow}

\usepackage{adjustbox}
\usepackage{varwidth}

\usepackage[accsupp]{axessibility}  

\usepackage{hyperref}

\usepackage{orcidlink}

\begin{document}

\title{Retrieval of sun-induced plant fluorescence in the O$_2$-A absorption band from DESIS imagery} 

\titlerunning{Retrieval of sun-induced fluorescence from DESIS imagery}

\author{Jim Buffat\inst{1}\orcidlink{0009-0006-8688-0580} \and
Miguel Pato\inst{2}\orcidlink{0000-0003-0111-0861} \and
Kevin Alonso\inst{3}\orcidlink{0000-0003-2469-8290
} \and Stefan Auer\inst{2}\orcidlink{0000-0001-9310-2337} \and Emiliano Carmona\inst{2}\orcidlink{0009-0008-8998-7310} \and Stefan Maier\inst{2}\orcidlink{0000-0001-6693-1973} \and Rupert Müller\inst{2}\orcidlink{0000-0002-3288-5814} \and Patrick Rademske\inst{1} \and \\ Uwe Rascher\inst{1}\orcidlink{0000-0002-9993-4588} \and Hanno Scharr\inst{4}\orcidlink{0000-0002-8555-6416}}

\authorrunning{J.~Buffat et al.}

\institute{Forschungszentrum Jülich GmbH, Institute of Bio- and Geosciences,\\ IBG-2: Plant Sciences, Jülich, Germany \and
Remote Sensing Technology Institute, German Aerospace Center
(DLR), Oberpfaffenhofen, Germany \and Starion Group c/o European Space Agency (ESA), Largo Galileo Galilei,  \\ Frascati 00044, Italy \and Forschungszentrum Jülich GmbH, Institute of Advanced Simulations, \\ IAS-8: Data Analytics and Machine Learning, Jülich,  Germany }

\maketitle

\begin{abstract}
  We provide the first method allowing to retrieve spaceborne SIF maps at 30 m ground resolution with a strong correlation ($r^2=0.6$) to high-quality airborne estimates of sun-induced fluorescence (SIF).
  SIF estimates can provide explanatory information for many tasks related to agricultural management and physiological studies. While SIF products from airborne platforms are accurate and spatially well resolved, the data acquisition of such products remains science-oriented and limited to temporally constrained campaigns. Spaceborne SIF products on the other hand are available globally with often sufficient revisit times. However, the spatial resolution of spaceborne SIF products is too small for agricultural applications. In view of ESA's upcoming FLEX mission
  we develop a method for SIF retrieval in the O$_2$-A band of hyperspectral DESIS imagery to provide first insights for spaceborne SIF retrieval at high spatial resolution. To this end, we train a simulation-based self-supervised network with a novel perturbation based regularizer 
  and test performance improvements under additional supervised regularization of atmospheric variable prediction.
  In a validation study with corresponding HyPlant derived SIF estimates at 740 nm we find that our model reaches a mean absolute difference of $0.78 \, \,  \mathrm{mW\,  nm^{-1} \,  sr^{-1} \, m^{-2}}$.
\end{abstract}

\section{Introduction}
\label{sec:intro}

The potential of sun-induced flurorescence (SIF) for agricultural management and phenotyping tasks was recognized early in the development of retrieval algorithms \cite{maierSunInducedFluorescenceNew2004}. 
Since SIF is fuelled by a residual energy flux of photosynthetically active radiation (PAR) that is not consumed by processes related to the plant's photochemistry and thermal energy dissipation it provides a causal link between radiance measurements and the photosynthetic status of plants \cite{meroniRemoteSensingSolarinduced2009,porcar-castellLinkingChlorophyllFluorescence2014,tolModelsFluorescencePhotosynthesis2014,verrelstGlobalSensitivityAnalysis2015}.
Various studies have utilized this relationship as the theoretical basis for stress detection and monitoring \cite{acMetaanalysisAssessingPotential2015,dammResponseTimesRemote2022,pintoDynamicsSunInduced2020,decanniereRemoteSensingInstantaneous2022,zhangSatelliteSolarinducedChlorophyll2023}, the estimation of photosynthetic activity and gross primary productivity \cite{chengIntegratingSolarInduced2013,tagliabueExploringSpatialRelationship2019,sunOverviewSolarInducedChlorophyll2018,zhangReductionStructuralImpacts2020}, crop monitoring and yield predictions \cite{guanImprovingMonitoringCrop2016,somkutiNewSpacebornePerspective2020,kiraScalableCropYield2024,pengAssessingBenefitSatellitebased2020} and disease monitoring \cite{calderonHighresolutionAirborneHyperspectral2013,rajiDetectionMosaicVirus2015} from SIF estimates derived from remote sensing data at various spatial scales.
Quantitative estimates of SIF allow for more sensitive and causally founded physiological assessments compared to purely reflectance based indices commonly used for such tasks.
Different studies have shown the increased explanatory power of SIF estimates measured at canopy level in a range of tasks \cite{dammResponseTimesRemote2022,oivukkamakiInvestigatingFoliarMacro2023,wangComparisonSatelliteDerived2024,liuImprovedVegetationPhotosynthetic2023}. 

SIF retrieval methods for a variety of sensors have been developed as the number of airborne and spaceborne sensors with sufficient spectral resolution has increased \cite{mohammedRemoteSensingSolarinduced2019}.
However, no spaceborne sensor designed specifically for fluorescence retrieval has yet been operationalized.
ESA's Earth Explorer Mission FLEX \cite{druschFLuorescenceEXplorerMission2017}, planned to be launched in 2025, will be the first such instrument.
Spaceborne SIF estimates to this day are derived from data acquired by satellite missions for atmospheric characterization (\eg, GOSAT \cite{joinerFirstObservationsGlobal2011}, GOME \cite{joinerGlobalMonitoringTerrestrial2013a, guanterGlobalTimeresolvedMonitoring2014}, SCIAMACHY \cite{joinerNewMethodsRetrieval2016}, 
OCO-2/3 \cite{sunOCO2AdvancesPhotosynthesis2017,elderingOCO3MissionMeasurement2019}, TROPOMI \cite{guanterPotentialTROPOsphericMonitoring2015, guanterTROPOSIFGlobalSuninduced2021}, TanSAT \cite{yaoNewGlobalSolarinduced2021}) as their spectral resolution (SR) and signal-to-noise ratio (SNR) allow for SIF retrieval from Fraunhofer lines \cite{frankenbergDisentanglingChlorophyllFluorescence2011,frankenbergChlorophyllFluorescenceRemote2012,druschFLuorescenceEXplorerMission2017}. 
However, the spatial resolution of these atmospheric sensors (> 4 km$^2$) is insufficient for most agricultural applications.
FLEX, on the other hand, will provide radiance data with a pixel size of 300 m which still imposes severe limits on its usability for a wide range of applications in heterogeneous agricultural landscapes.

As an exploratory step towards spaceborne SIF retrieval at high spatial resolution, we therefore propose a deep learning architecture and a loss formulation for the first SIF retrieval from hyperspectral imagery of the DLR Earth Sensing Imaging Spectrometer (DESIS).
SIF retrieval from DESIS imagery has the benefit of providing spaceborne SIF products at an unprecedented spatial resolution of 30 m which principally allows for the targeted acquisition of auxiliary validation data at high spatial resolution for the upcoming FLEX mission.
However, the SR and SNR of DESIS are insufficient for consistent SIF retrieval with current traditional retrieval methods leveraging data in the O$_2$-A absorption band \cite{fletcherReportMissionSelection2015,dammFLDbasedRetrievalSuninduced2014,liuEffectsSpectralResolution2015} where the fluorescence signal contribution to the at-sensor signal has a local maximum.
Airborne SIF retrieval with similar methods applied to radiance data at lower SR has however been shown to yield consistent relative SIF estimates \cite{belwalkarEvaluationSIFRetrievals2022}.
As a solution, we extend the simulation-based self-supervised deep learning approach of \cite{buffatDeepLearningBased2023,buffatMultiLayerPerceptron2024}, called Spectral Fitting Method Neural Network (SFMNN), originally developed with airborne hyperspectral imagery.
As in other self-supervised simulation-based learning schemes, this approach leverages the implicit constraints of a differentiable simulator of the physical image generation in the loss \cite{hendersonLearningGenerateReconstruct2018,jatavallabhulaGradSimDifferentiableSimulation2021} and primarily does not rely on labels for training.
Further regularization terms that enforce physical and physiological domain constraints allow this encoder-decoder architecture to decompose and reconstruct hyperspectral data in the spectral range around the O$_2$-A absorption band.

In this contribution we introduce regularization terms in the SFMNN framework allowing consistent SIF retrieval in DESIS imagery despite its lower SNR and SR.
Firstly, we propose a perturbation based augmentation scheme to promote the decorrelation of the predicted SIF from other confounding variables affecting the at-sensor signal.
Secondly, we show that including ancillary atmospheric data from DESIS L2A products by means of a secondary supervised downstream learning task improves the performance of our model.

\section{Data}

\subsection{DESIS observation, simulation and emulation}
\begin{figure}[t]
    \centering
    \includegraphics[width=1\textwidth]{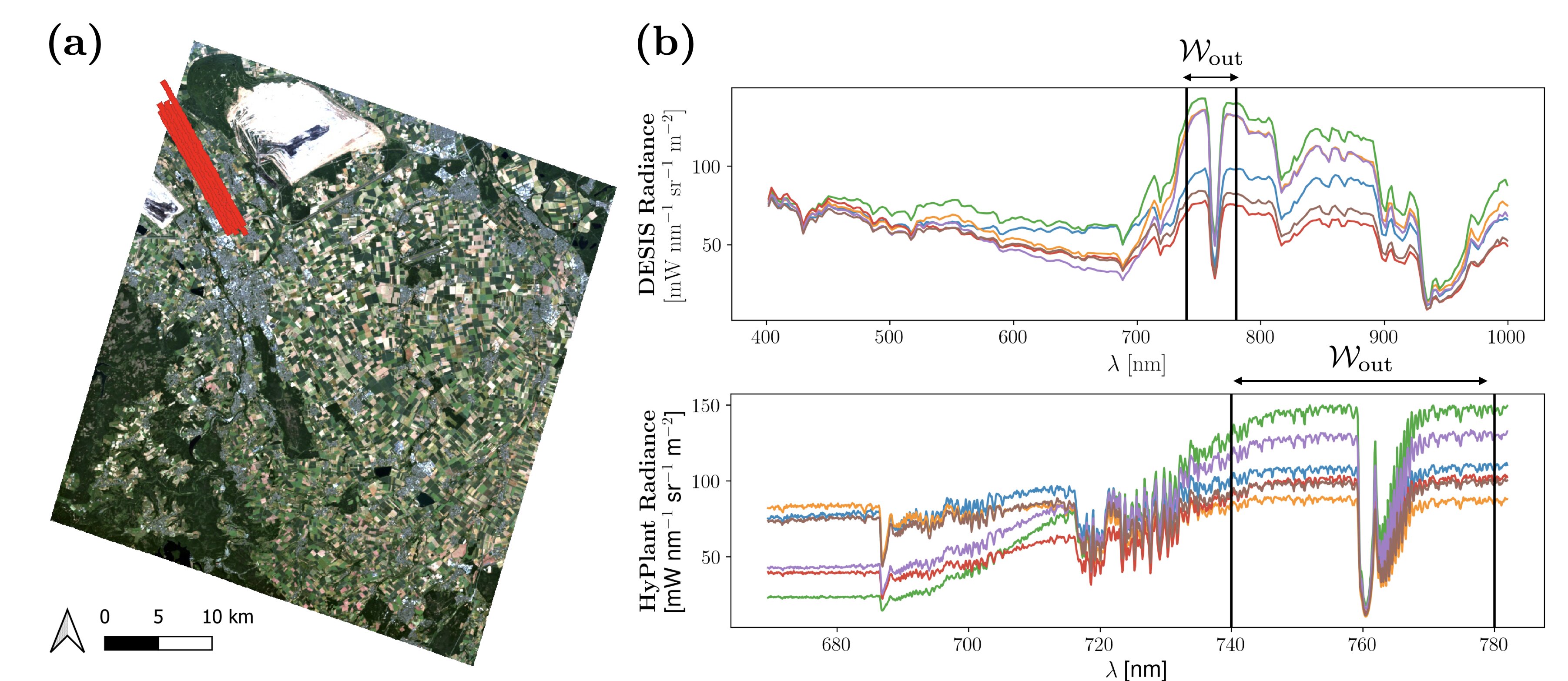}
        \caption{
DESIS and HyPlant data. \textbf{(a)} RGB composite of a DESIS acquisition (13/06/2023 14:37 CEST) and, in red, extent of spatially and temporally overlapping HyPlant acquisitions (13/06/2023 14:11 - 14:38 CEST). \textbf{(b) }Top: Sample DESIS at-sensor radiance spectra, Bottom: 
sample HyPlant at-sensor radiance spectra. $\mathcal{W}_\mathrm{out}$ denotes the spectral emulator domain.}
        \label{fig:data_overview}
\end{figure}

The DLR Earth Sensing Imaging Spectrometer (DESIS) is a hyperspectral imaging sensor onboard the International Space Station (ISS) \cite{krutzInstrumentDesignDLR2019}.
It measures at-sensor radiance in 235 bands in the spectral range from 400 nm to 1000 nm with a nominal spectral sampling interval (SSI) of 2.55 nm and a full width at half maximum (FWHM) of 3.55 nm.
The spatial dimensions of DESIS acquisitions are fixed to 1024 $\times$ 1024 pixels with a nominal pixel width of 30 m.
See \cite{alonsoDataProductsQuality2019} for a complete description of the DESIS sensor, data products and associated uncertainties. 
We make use of 
a polynomial emulator formulation \cite{patoFastMachineLearning2023, patoPhysicsbasedMachineLearning2024} to reconstruct DESIS radiance spectra around the O$_2$-A absorption band.
We found small domain shift errors with respect to smile-corrected L1B DESIS products (see \cref{sec:results_general}). 
For training, we assembled a data set of 96 DESIS data takes (L1B and L2A in sensor geometry) matching either \mbox{OCO-3} or HyPlant recordings \cite{hyplant_desis_oco3_dataset}.  
The georegistration of DESIS SIF estimates was conducted with the operationally provided DESIS L2A geolayers.


\begin{table}[t]
    \centering
    \caption{\label{tab:data} Training (\textit{Trn.}) and validation (\textit{Val.}) data sets. $N_\text{px}$: number of covered DESIS pixels,  $N_\text{acq}$: number of DESIS acquisitions (with matching pixels, in the case of data sets \textit{HyPlant} and \textit{\mbox{OCO-3}}). A complete account of the data set is available \cite{hyplant_desis_oco3_dataset}.}
    \begin{tabular*}{\linewidth}{@{\extracolsep{\fill}} lllrrr}
    \textbf{Data Set} & $N_\text{px}$ &  $N_\text{acq}$ & Location & Method  & Type \\
    \toprule
    HyPlant &   10'196 (2020), 18'850 (2023) & 4 & Jülich (Germany) & \cite{cogliatiSpectralFittingAlgorithm2019} & Val. \\
    \midrule 
    \mbox{OCO-3} &   670 & 92 & Global & 
    \cite{sunOverviewSolarInducedChlorophyll2018} & Val. \\
    \midrule\midrule
    DESIS & $100 \times 10^6$ & 96 & Global & 
    --\phantom{[]}  & Trn. \\
    \bottomrule
    \end{tabular*}
\end{table}

\subsection{HyPlant campaigns 2020 and 2023}
HyPlant is an airborne spectrometer system providing hyperspectral radiance measurements with an SSI of 0.11 nm and a nominal FWHM of 0.25 nm \cite{siegmannHighPerformanceAirborneImaging2019}.
It is the airborne demonstrator version for FLEX \cite{druschFLuorescenceEXplorerMission2017} which is first spaceborne sensor specifically designed for SIF retrieval. 
As a result, HyPlant measurements have been used for SIF retrieval in yearly field campaigns since 2014 \cite{rascheruweTechnicalAssistanceDeployment2017a,rascheruweTechnicalAssistanceDeployment2017,rascheruweTechnicalAssistanceDeployment2018,rascheruwePhotoproxyTechnicalAssistance2019,rascheruweFLEXSentinelTandem2021,rascheruweFLEXSenseTechnicalAssistance2022,rascheruweHyPlantFLEXSimulator2022}.
For this contribution, spatially overlapping acquisitions of DESIS and HyPlant could be recorded on 13/06/2023.
Six HyPlant and two DESIS acquisitions were acquired within small time intervals of 1 - 25 minutes at around 14:30 CEST (cf. \cref{fig:data_overview} and \cref{tab:data}).
Additionally, we found close spatial matches between six HyPlant and two DESIS acquisitions on 23/06/2020 in the same region. In this case the HyPlant acquisitions were recorded at least an hour earlier than the DESIS acquisitions (12:08 CEST). 
This unique disposition of spatially and temporally matching spaceborne and airborne radiance measurements allowed us to compile a comparative data set of georegistered HyPlant and DESIS SIF estimates.
To this end, we processed the HyPlant at-sensor radiance with the Spectral Fitting Method (SFM) \cite{cogliatiSpectralFittingAlgorithm2019} to derive high-quality SIF estimates.
The alignment of DESIS and HyPlant SIF products involved downscaling (isotropic Gaussian smoothing and spatial resampling) HyPlant SIF to DESIS resolution.

\subsection{\mbox{OCO-3} SIF estimates}
\mbox{OCO-3} is a spectrometer assembly originally designed for the retrieval of column carbon dioxide \cite{elderingOCO3MissionMeasurement2019}. 
As DESIS, \mbox{OCO-3} is located onboard the ISS.
The high SR of the radiance measurements around the O$_2$-A absorption band of this sensor allows for SIF retrieval in this spectral region similarly to earlier spaceborne sensors designed for the retrieval of atmospheric gas compositions \cite{taylorOCO3EarlyMission2020}.
Since both \mbox{OCO-3} and DESIS are 
on the ISS, there exists a set of overlapping acquisitions with small time differences (< 10 minutes). 
We have identified a set of approximately 100 DESIS acquisitions that are partially covered by \mbox{OCO-3} measurements, exhibit a low ratio of cloud cover and are flagged to be of acceptable quality.
We make use of an \mbox{OCO-3} SIF product of those acquisitions \cite{oco3_data_2021,doughtyGlobalGOSATOCO22022} as a complementary performance validation of our DESIS SIF estimates.
These \mbox{OCO-3} SIF estimates were compared to DESIS pixels in a 300 m radius around the center of individual soundings.

\section{Architecture and Simulation-Based Loss}
\label{sec:methods}
\subsection{Architecture}
\begin{figure}[t]
    \centering
    \includegraphics[width=1\textwidth]{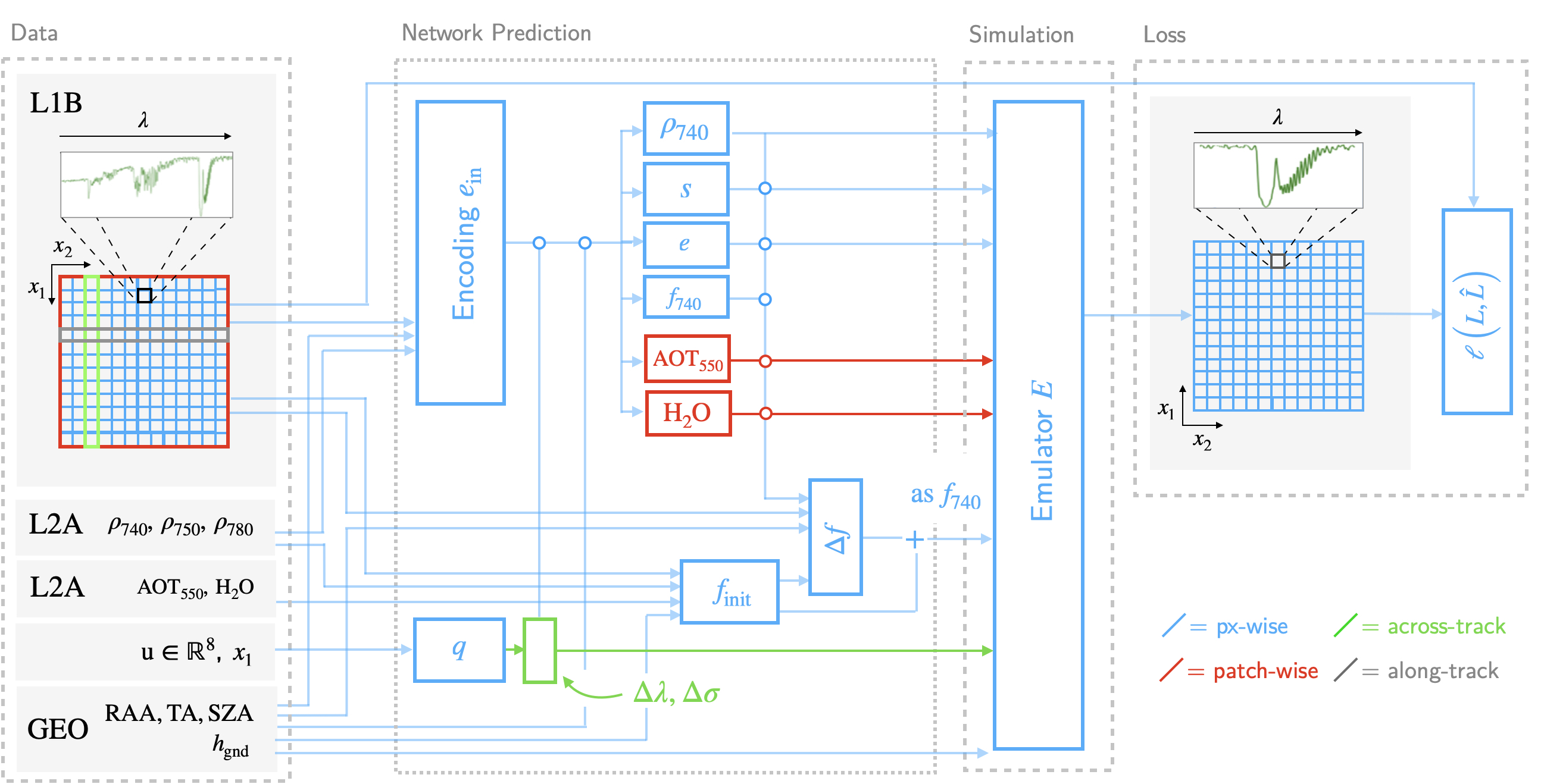}
        \caption{Proposed network architecture. \textbf{Data}: gray blocks denote different data sources: \textit{L1B} smile-corrected DESIS L1B at-sensor radiance, \textit{L2A} reflectance and atmospheric variables provided in the DESIS L2A product, \textit{GEO} geometrical variables from L1C metadata and L2A geolayer: RAA (relative azimuth angle), TA (tilt angle), SZA (sun zenith angle), $h_\mathrm{gnd}$ (digital elevation model). \textit{other}: $u$ denotes trainable sensor state identifier and $x_1$ the across-track pixel position.
        \textbf{Network}: variables ($\rho_{740}$, $s$, $e$, $f_{740}$) predicted by $d_\mathrm{px}$ and ($\mathrm{AOT}_{550}$, H$_2$O) predicted by $d_\mathrm{patch}$ as well as ($\Delta\lambda$, $\Delta\sigma$) predicted by $q$ are passed to the simulation layer implemented as the emulator $E$ \cite{patoFastMachineLearning2023,patoPhysicsbasedMachineLearning2024}.}
        \label{fig:architecture}
\end{figure}
 
The SIF retrieval method for DESIS imagery presented in this work is based on the Spectral Fitting Method Neural Network (SFMNN) \cite{buffatDeepLearningBased2023,buffatMultiLayerPerceptron2024}.
This network implements in an encoder-decoder type architecture to fit parameters $p_j$ of a simulation model of observational at-sensor radiance data.
The simulation model parameterizes the physical signal generation as a function of surface, atmospheric, sensor and geometrical variables.
As a result, SIF retrieval is formulated as a feature optimization for optimal spectral decomposition and reconstruction.
In order to constrain the solution space, the output dimensionality of the prediction of the simulation parameters is variable (\cref{fig:architecture}). 
While surface parameters are allowed to vary in a pixelwise fashion, the atmospheric parameters are constrained to a single scalar value for pixels in a single input patch, i.e., within the same spatial neighbourhood. 
This is motivated by the large spatial auto-correlation distance of the atmospheric variables which is typically larger than the patches of 30 $\times$ 30 DESIS pixels (900 $\times$ 900 m) used during training.
Equally, simulated sensor miscalibration 
only varies along the across-track dimension.

We implement an encoder, decoders $d_\text{px}$ and $d_\text{patch}$ for the surface and atmospheric variables and a module $q$ for the sensor variables.
The module architecture consists of stacked multi-layer perceptrons (MLPs) with residual links (see \cite{buffatMultiLayerPerceptron2024} for a detailed architecture description).
The simulation model implemented here exhibits three major differences with respect to \cite{buffatDeepLearningBased2023,buffatMultiLayerPerceptron2024} that allow the application of SFMNN in an adapted form to DESIS data (explained below).

\subsubsection{Emulator}
We replace the approximate simulation model of \cite{buffatDeepLearningBased2023,buffatMultiLayerPerceptron2024} with an emulator of a simulations of DESIS-like radiance spectra around the O$_2$-A absorption band. 
To this end, we adopt a simulation tool and emulation set-up introduced by Pato \etal \cite{patoFastMachineLearning2023,patoPhysicsbasedMachineLearning2024} who show that
a polynomial emulator of 4$^{th}$ degree yields an approximation error that is significantly smaller than typical at-sensor fluorescence in a DESIS-like configuration.
Polynomial emulators are not widely used for radiance emulation in remote sensing (\eg \cite{verrelstEmulationLeafCanopy2016,verrelstSCOPEBasedEmulatorsFast2017}). 
In our specific case, only the small spectral range $\mathcal{W}_\mathrm{out} = $[740 - 780] nm around the O$_2$-A band must be covered, however, such that a model with small complexity is able to meet the precision requirements.
The polynomial nature of the emulator is advantageous since (i) it can be integrated easily in a feed-forward neural network architecture as a fixed linear layer and (ii) it is computationally efficient such that training and prediction are not significantly affected by it.

\subsubsection{Residual Fluorescence Estimation}
SFMNN is a completely self-supervised approach that does not require any labelled data to be trained to a set of hyperspectral imagery. 
Preliminary tests with a plain SFMNN approach on DESIS data did not provide SIF estimates with useful sensitivity to the HyPlant and \mbox{OCO-3} validation data, however.
The most likely cause for this is the low SR of DESIS data, especially in comparison to the HyPlant data on which SFMNN was originally developed. 
We therefore adopt a modification to SFMNN whereby the fluorescence $f$ is not estimated directly from radiance data.
Instead, a residual $\Delta f$ to an initial guess $f_\mathrm{init}$ with large uncertainty is predicted by a dedicated module from L1C and L2A data (cf. \cref{fig:architecture}) such that we can interpret
\begin{equation}
    f_{740} = f_\mathrm{init} + \Delta f,
\end{equation}
as the model's SIF estimate. 
A similar approach has notably be implemented by Brodrick \etal \cite{brodrickGeneralizedRadiativeTransfer2021} to improve a low-quality atmospheric radiative transfer model.
$f_\mathrm{init}$ here denotes an initial estimate of the fluorescence emission that we gain from a supervised predictor directly trained on simulated DESIS data. 
While the simulation-trained model alone yields noisy predictions with subpar performance (cf. \cref{fig:prediction_example}), we find that its combination with the self-supervised principles of SFMNN  results in significantly improved prediction accuracy.

\begin{figure}[t]
    \centering
    \includegraphics[width=1\textwidth]{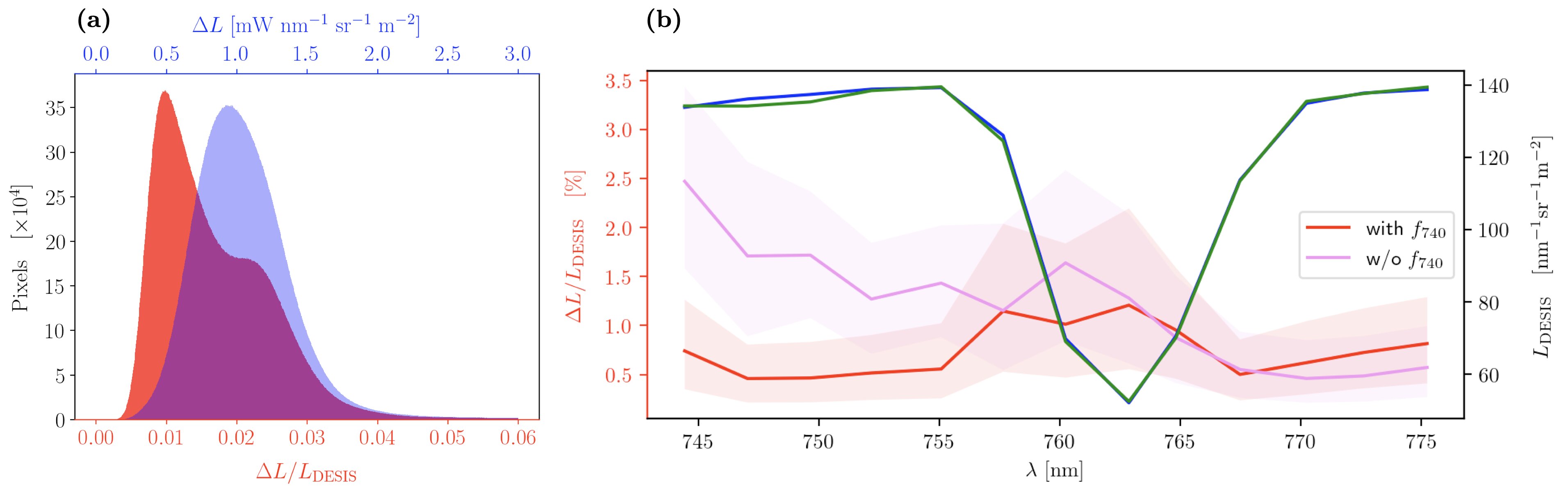}
        \caption{\textbf{(a)} Relative and absolute reconstruction error of best performing model configuration over all DESIS acquisitions. \textbf{(b)}\textit{ Red/Pink: }Spectrally explicit error distribution in the DESIS acquisition matching the \mbox{OCO-3} validation data (\cref{fig:data_overview}), light colors denote the 25 - 75 percentiles. \textit{Blue/Green}: Sample reconstruction (blue) of a single spectral DESIS observation (green) matching HyPlant (2023) data.}
        \label{fig:err_per_wvl}
\end{figure}

\subsubsection{Sensor state}
Miscalibration of a hyperspectral sensor can be expressed in terms of changes to a Gaussian approximation of the instrument spectral response function (ISRF) in each spatio-spectral pixel. 
Commonly, such miscalibrations are parameterized with shifts of its standard deviation $\Delta \sigma$ and shifts of its center wavelength $\Delta \lambda$.
Hyperspectral sensors can suffer from changing $\Delta \lambda$ and $\Delta \sigma$ due to mechanical and environmental stresses that change the ISRF and must be addressed operationally by periodical recalibration.
\begin{table}[b!]
    \centering
     \caption{\label{tab:architecture} Dimensions $e_\mathrm{in}$, $d_\text{px}$, $d_\text{patch}$, $q$ and $\Delta f$ (cf. Fig. \ref{fig:architecture}). Modules consist of stacked MLPs. Each element in \textit{Dim.} denotes the dimension of individual perceptrons in a single MLP, \textit{Reps.} the number of perceptrons in a single MLP (all with the same dimension reported in \textit{Dim.}) and $D_p$ the dropout rate of the output of each MLP. For a detailed architecture description see \cite{buffatMultiLayerPerceptron2024}.}
        \begin{tabular*}{\linewidth}{@{\extracolsep{\fill}}lrr}
        Module & Parameters & \\
        \toprule
        \multirow{3}{*}{Encoder $e_\mathrm{in}$}      &  \textit{Dim.} & \textsf{(1000, 500, 200, 100, 50, 50, 50, 30)} \\
            & \textit{Reps.}& \textsf{(2, 3, 3, 3, 3, 3, 3, 3)} \\
            & $D_p$ & \textsf{(0.05, 0.01, 0.01, 0.01, 0.005, 0.001)} \\
        \midrule
        \multirow{3}{*}{Decoders $d_\text{px}$, $d_\text{patch}$ and $q$}      &  \textit{Dim.} & \textsf{(100, 50, 50, 50)} \\
          & \textit{Reps.}& \textsf{(3, 2, 2, 2)} \\
            & $D_p$ & \textsf{(0.001, 0.001, 0.0)} \\
        \midrule
        \multirow{3}{*}{\makebox[0pt][l]{$\Delta f$}}      &  \textit{Dim.} & \textsf{(1000, 200, 100, 50, 50, 50)} \\
            & \textit{Reps.}& \textsf{(2, 3, 3, 3, 3, 1)} \\
            & $D_p$ & \textsf{(0.05, 0.005, 0.001)} \\
        \midrule
        \end{tabular*}
\end{table}
The issue of mismatching calibrations may be significant due to (i) DESIS' overall low SR and (ii) a subpar smile-correction of L1C in the O$_2$-A band.
In order to alleviate this issue the shifts $\Delta \sigma$ and $\Delta \lambda$ are fitted as a function $q$ of acquisition-specific identifiers $u$ such that sensor drifts, processing changes and artefacts may be accounted for.
In order to prevent a too large degree of freedom that could affect the SIF prediction, we implement a module $q$ as a decoder module with only $u$ and sensor position $x$ as inputs such that (i) it is independent of any other input than the identifier $u$, (ii) multiple acquisitions may have the same identifier (acquisitions of the same date have the same identifier $u$) and (iii) only across-track variability of the shifts are allowed as is realistic for a push-broom sensor.

\subsection{Loss formulation}

We propose a simulation-based loss 
$\ell  = \ell_\mathrm{res} + \ell_m + \ell_{\Delta f} + \ell_N + \ell_c$
where $\ell_\mathrm{res}$ evaluates the reconstruction residuals of the model with respect to the observational input, $\ell_m$ and $\ell_{\Delta f}$ ensure that the network matches prior knowledge, $\ell_N$ ensures the physiological plausibility of the SIF estimates and $\ell_c$ denotes a perturbation based regularization that enhances the decorrelation between predicted variables by means of a physically accurate augmentation. 
Self-supervised learning with radiance observations is addressed by adopting the methodology of \cite{buffatDeepLearningBased2023,buffatMultiLayerPerceptron2024}, where the reconstructed signal is compared to the observation similarly to other self-supervised methods such as masked auto-encoders \cite{heMaskedAutoencodersAre2021,hongSpectralGPTSpectralRemote2024a}.
A squared residual over the whole spectrum as well as a weighted residual boosting the loss in spectral regions with high average fluorescence contributions punish the network for not reproducing  the at-sensor observations.
This is implemented by
\begin{equation}
    \ell_\mathrm{res}\left(L, \hat L\right) = \left\langle\left(L - \hat L\right)^2\right\rangle_{\lambda,\, x} +
    \frac{\gamma_f}{|\mathcal{W}_\mathrm{out}|} \left\langle  \sum_{\mathclap{\quad \,\,\,\lambda \in \mathcal{W}_\mathrm{out}}} w_\lambda \left(L(\lambda) - \hat L( \lambda)\right)^2 \right\rangle_{x}, 
\end{equation}
where $\langle \dots \rangle_{\lambda,x}$ denotes the batchwise mean over the spatial and spectral dimension, and where $ L$ and $\hat L$ denote the observations and emulated predictions in the spectral range $\mathcal{W}_\mathrm{out}$. 
The weighting $w_\lambda$ is resampled from the weights originally proposed in \cite{buffatDeepLearningBased2023,buffatMultiLayerPerceptron2024} for the specific sensor characterization of the DESIS sensor.
Furthermore, we also adopt the selective gradient backpropagation of the second term which is set to only affect the fluorescence prediction $\hat f_{740}$.

The inclusion of prior information on atmospheric variables and the SIF emission are implemented as regularization terms 
\begin{equation}
    \label{eq:ell_m}\ell_\mathrm{m} = \sum_{k \in \mathcal{K}} \gamma_{p_k} \left(p_k' - \hat p_k \right)^2 \quad \mathrm{and} \quad \ell_{\Delta f} = \gamma_{\Delta f} \left(\hat f - f_\mathrm{init} \right)^2
\end{equation}
where $f_\mathrm{init}$ denotes the prediction of the supervised SIF predictor and  where $\mathcal{K} = \{\mathrm{H_2O}, \, \mathrm{AOT_{550}}\}$.
We denote by H$_2$O the water vapour density and by AOT$_{550}$ the aerosol optical thickness at 550 nm, which are derived operationally and distributed with DESIS L2A products. 
Thus, $\ell_m$ introduces a supervised regression training of ancillary data from DESIS products as a secondary task.
As an alternative strategy, we test in \cref{sec:results_lAOT} a set-up where the ancillary data is passed directly to the input and the emulator for training and prediction.
Accordingly, our model does not provide predictions of the atmospheric parameters (H$_2$O, AOT$_\mathrm{550}$) in this latter variant.

The SIF estimate $f_\mathrm{init}$, which we gain from a predictor trained on simulated DESIS data, is included in the loss parallelly to $\ell_m$. 
This effectively controls the range of deviation that the residual module $\Delta f$ is allowed to introduce.
Additionally, the fluorescence estimate of the residual module is controlled by the constraint
\begin{equation}
    \ell_N = \gamma_N \, \hat f \, \delta\left(\mathrm{NDVI}(L) \leq \tau\right)
\end{equation}
to ensure vanishing SIF predictions in pixels without vegetation, i.e., with small NDVI \cite{buffatDeepLearningBased2023,buffatMultiLayerPerceptron2024}.
We fix $\tau = 0.15$ in all experiments.

Finally, we leverage the compact physical description of the generation of at-sensor radiance given by the emulator to derive a perturbation based augmentation for regularization that we will refer to as \textit{consistency regularization}.
\begin{figure}[t]
    \centering
    \includegraphics[width=1\textwidth]{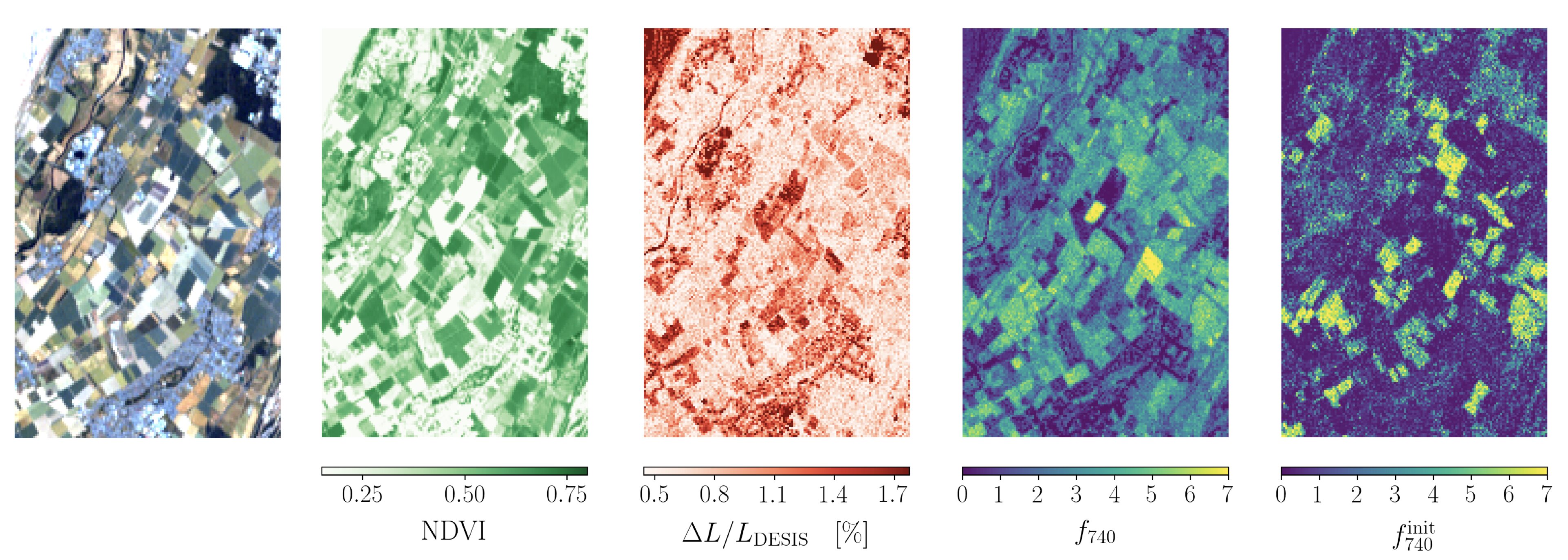}
        \caption{Overview of an image excerpt of a DESIS acquisition matching HyPlant (2023) validation data (\cref{fig:err_per_wvl}). Left to right: RGB composite, NDVI derived from L2A, relative reconstruction error, fluorescence estimate $f_{740}$, initial fluorescence estimate $f^\mathrm{init}_{740}$.}
        \label{fig:prediction_example}
\end{figure}
We denote by $g_{p_j}$ the prediction of emulator input variables $p_j$.
Consequently, $g_{p_j}$ are approximate partial inversions of the emulator $e$. 
Perturbations of the input radiance $L$ by $\delta L$ affect the predictor $g_f$ and vice-versa perturbations $\delta f$ of the predicted fluorescence $\hat f$ affect the emulator:
\begin{equation}
    g_f(L + \delta L) = \hat f + \delta f \, \, \,  \mathrm{and} \, \, \,  e(\hat f + \delta f,\, p_j) = L_{e}\ + \delta L_e
\end{equation}
where the spectral range $\mathcal{W}_\mathrm{out}$ of the emulator output $L_e$ is smaller than the range of the observational DESIS spectrum $L$ due to a practical limitation of the emulator design. 
We can simulate new approximate samples by generating radiance perturbations $\delta L_e$.
To do this, we write
\begin{align}
    (L + \delta L)(\lambda) \approx  L'_{\delta f} &= L(\lambda) + \delta L_e \, \,\delta\left(\lambda \in  \mathcal{W}_\mathrm{out}\right), \\
    &\text{where} \,\,\, \delta L_e = e(\hat f + \delta f, p_j) - e(\hat f, \, p_j). \nonumber
\end{align}
Neglecting any simulation errors and inaccuracies due to the perturbation implementation of $L'_{\delta f}$ an optimal solution should yield
\begin{equation}
    g_f\left(L'_{\delta f}\right) - \hat f = \delta f \, \, \, \mathrm{and} \,\,\,  g_{p_j}\left(L'_{\delta f}\right) = p_j,
\end{equation}
since all changes in the perturbed observation can be attributed to a change in fluorescence in this case.
We thus formulate the regularization as 
\begin{equation}
    \ell_c(L, \hat f, \hat p_j) = \mathbb{E}_{\delta f \sim \mathcal{F}}\left[\left(g_f\left(L'_{\delta f}\right)  - (\hat f + \delta f) \right)^2 + \left(g_{p_j}\left(L'_{\delta f}\right) - \hat p_j\right)^2\right]
\end{equation}
where $\mathcal{F}$ is the fluorescence range over which $e$ is defined.
We implement the expectation as a mean over a uniformly sampled set of $\delta f$  in each training forward pass.

\begin{figure}[t]
    \centering
    \includegraphics[width=0.97\textwidth]{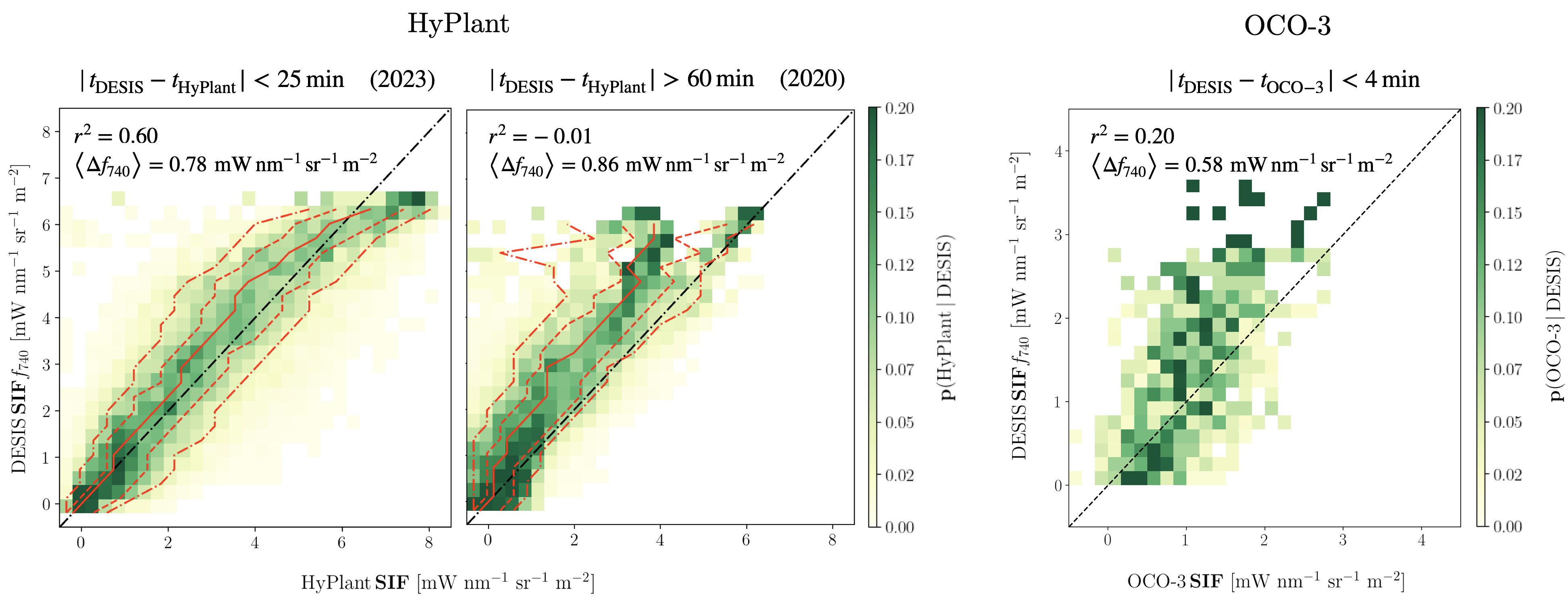}
        \caption{Conditional 2d-histogram of DESIS SIF estimates of the best performing run \mbox{($\gamma_\mathrm{c} = 5 \times 10^3$, $\gamma_\mathrm{AOT} = 100$)} compared to HyPlant (2023), HyPlant (2020) and \mbox{OCO-3} validation data sets \cref{tab:data}, red dashed lines denote the 10, 25, 75 and 90 percentiles.}
        \label{fig:hyplant_oco3_comparison}
\end{figure}

\section{Results}

\subsection{Reconstruction Performance \& Validation}
\label{sec:results_general}
The network must reconstruct observed spectra with high accuracy.
The model trained on all DESIS acquisitions with matching HyPlant or \mbox{OCO-3} SIF products (\cref{tab:data}) with the proposed loss (with $\gamma_f = 1$, $\gamma_{\Delta f} = 5$, $\gamma_\mathrm{H_2O} = 1$ and $\gamma_N = 10$ fixed during preliminary experimental runs and $\gamma_\mathrm{AOT} = 100$ and $\gamma_c = 5 \times 10^3$ being the best configuration) 
reaches a mean relative reconstruction error of 1.6 \% (\cref{fig:err_per_wvl} (a)).
The spectral variation of the reconstruction error is small (\cref{fig:err_per_wvl} (b)) which evidences the model's capacity to plausible signal generation across the spectral domain. 
We equally observe that the reconstruction error is reduced under inclusion of $f_{740}$ compared to reconstructions where we fixed $f_{740} = 0$. 

In \cref{fig:prediction_example} we show an exemplary result.
The prediction of $f_{740}$ exhibits a reduced noise level compared to $f_{740}^\mathrm{init}$ and correlates with the distribution of agricultural fields. 
In order to assess the performance quantitatively we compare DESIS SIF estimates to matching HyPlant and \mbox{OCO-3} SIF estimates (\cref{tab:data}).
Since existing SIF retrieval methods are not adapted to DESIS data (cf. \cref{sec:methods}) we rely on these estimates as ground truth.
The same configuration as above yields a mean absolute difference $\langle\Delta f_{740}\rangle_\mathrm{HyP23} = 0.78 \, \,  \mathrm{mW\,  nm^{-1} \,  sr^{-1} \, m^{-2}}$ (smaller is better) and a coefficient of determination $r^2_\mathrm{HyP23} = 0.6$ (larger is better) in the HyPlant (2023) data set.
The DESIS estimates perform worse in a comparison with HyPlant (2020) data ($\langle\Delta f_{740}\rangle_\mathrm{HyP20} = 0.86 \, \,  \mathrm{mW\,  nm^{-1} \,  sr^{-1} \, m^{-2}}$  and $r^2_\mathrm{HyP20} = -0.01$) due to a large overestimation of our approach.
This is expected since the data was recorded closer to solar noon when the diurnal course of SIF peaks \cite{wangDiurnalVariationSuninduced2021,changUnpackingDriversDiurnal2021}.
Finally, we find
$\langle\Delta f_{740}\rangle_\mathrm{OCO3} = 0.58 \, \,  \mathrm{mW\,  nm^{-1} \,  sr^{-1} \, m^{-2}} $ and $r^2_\mathrm{OCO3} = 0.2 $ compared to \mbox{OCO-3} data (cf. \cref{fig:hyplant_oco3_comparison}).

\begin{figure}[b!]
    \centering
    \includegraphics[width=0.9\textwidth]{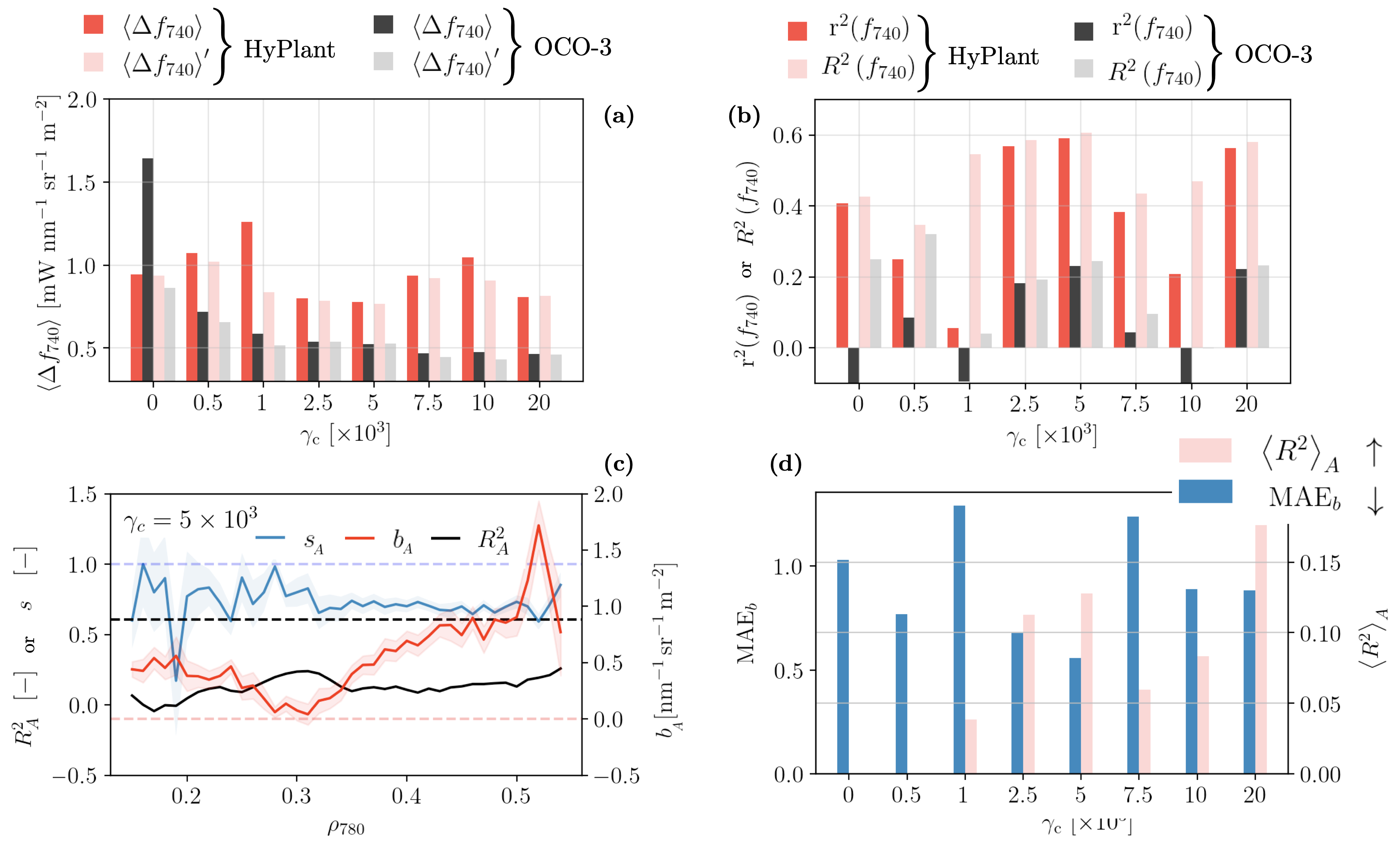}
        \caption{\textbf{(a)} and \textbf{(b)}: performance of the model trained with varying $\gamma_c$ as compared to HyPlant (2023) and \mbox{OCO-3} SIF. \textbf{(c)}: Reflectance constrained metrics for the case $\gamma_\mathrm{AOT} = 0$, $\gamma_c = 5\times 10^3$, light blue and light red colors denote the uncertainty of the least-squares fit to gain $s$ and $b$, \textbf{(d)}: metrics under varying $\gamma_c$ in HyPlant (2023) data.}
        \label{fig:gamma_c_comparison}
\end{figure}

We evaluated the consistency regularization and the inclusion of ancillary data on the SIF prediction. 
We only validated with respect to the HyPlant (2023) and \mbox{OCO-3} data sets as the acquisition time difference of HyPlant (2020) would have introduced large biases.
In addition to the metrics above, we report $\langle\Delta f_{740}\rangle$ and $r^2$ 
for bias corrected data to differentiate between performance increases due to bias reduction and due to increased explanation of label variance, i.e., 
\begin{equation}
    R^2 = r^2\left(n \circ f_{740}, n \circ \hat f_{740} \right) \, \, \mathrm{and} \, \, \left\langle \Delta f_{740}\right\rangle' = \left\langle \Delta\left(n\circ f_{740}\right)\right\rangle,
\end{equation}
where $n(x) = x - \langle x\rangle$.
Furthermore, since there is an empirical correlation between SIF and reflectance due to common causal drivers, a stronger validation consists in conditioning the model's performance on reflectance $\rho$.
Subsequently, we define the \textit{reflectance constrained explained variation}
\begin{equation}
    \label{eq:r2_reflc}\left\langle R^2\right\rangle_{A} = |P|^{-1}\sum_{\rho_{780} \in P} R^2(A_{\rho_{780}}), \quad  A_{\rho_{780}} = p\left(f_{740} \, \Big| \, \hat{f}_{740}, \, |\rho -\rho_{780}| < d\rho\right)
\end{equation}
where $d\rho = 0.02$ and exclusively reflectance in a single DESIS band ($\rho_{780}$) is considered.
Only the HyPlant (2023) data set is large enough to calculate $\left\langle R^2\right\rangle_{A}$ confidently, however.
We focus on the reflectance at 780 nm since the influence of SIF on DESIS L2A reflectances at 780 nm is negligible while $\rho_{780}$ also is strongly correlated to vegetation cover.
Similarly, we characterize $A_{\rho_{780}}$ by the slope $s_{A}$ and bias $b_{A}$ of a linear model fitted to it (cf. \cref{fig:gamma_c_comparison} (c)).
Since we expect this bias to vanish for perfect predictions we calculate the \textit{mean reflectance dependent bias} $\mathrm{MAE}_b = \langle b_{A} \rangle $ as an additional performance metric.
 
  
\subsection{Consistency Regularization Weighting $\gamma_c$}
\label{sec:results_lC}

In order to evaluate the impact of the consistency regularization $\ell_c$ on DESIS SIF, we perform a grid search over $\gamma_c$ (\cref{fig:gamma_c_comparison} (a) and (b)) without including $\ell_m$ in the training loss.
In the \mbox{OCO-3} data set the inclusion of  $\ell_c$ at all tested weights $\gamma_c$ outperforms the case $\gamma_c = 0$ in terms of $\langle \Delta f_{740}\rangle$ (smaller is better) and generally also in terms of $R^2$ (larger is better) (Figs. \ref{fig:gamma_c_comparison} (a) and (b)).
With HyPlant (2023) we find only a localized performance optimum at $\gamma_c \sim 5 \times 10^3$ in terms of $\langle \Delta f_{740}\rangle$ and $r^2$.
We show in \mbox{\cref{fig:gamma_c_comparison} (c)} that we find particularly strong overestimation at $\rho_{780}>0.5$
Improved performance under $\ell_c$ in HyPlant, however, can be seen in terms of $\mathrm{MAE}_b$ and $\left\langle R^2\right\rangle_{A}$ (\cref{fig:gamma_c_comparison} (d)).
Specifically, we find $\left\langle R^2\right\rangle_{A} > 0$ only if the consistency regularization is applied.


\subsection{Inclusion of Ancillary Data}
\label{sec:results_lAOT}

\begin{figure}[t]
    \centering
    \includegraphics[width=1\textwidth]{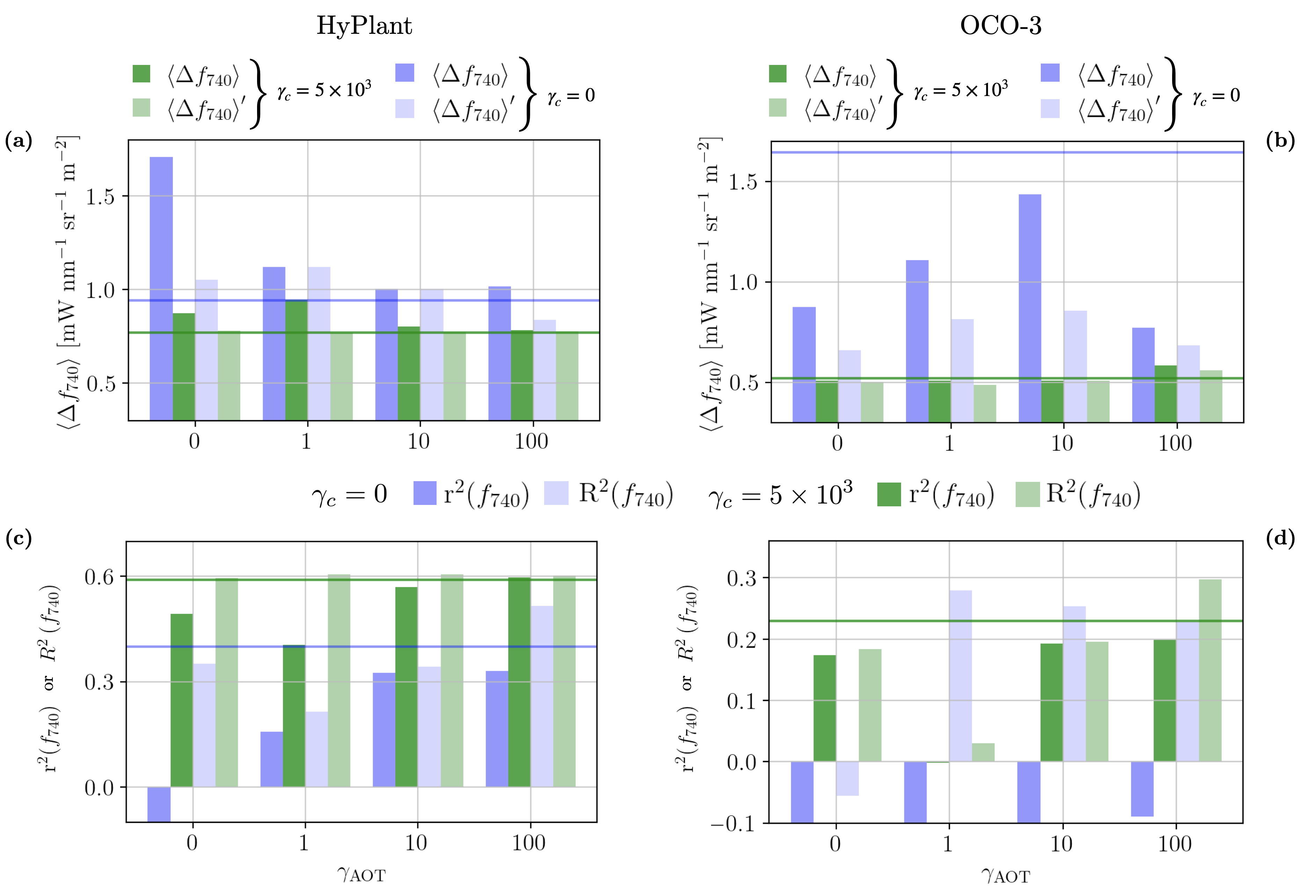}
        \caption{Model performance with respect to HyPlant (2023) SIF in \textbf{(a)} and \textbf{(c)} and \mbox{OCO-3} SIF in \textbf{(b)} and \textbf{(d)} under variable $\gamma_\mathrm{AOT}$. Light colors indicate  $\gamma_c = 0$, dark colors indicate $\gamma_c = 5 \times 10^3$. Horizontal lines indicate the performance of the model runs without ancillary data in \cref{sec:results_lC}.}
        \label{fig:gamma_aot}
\end{figure}

\begin{figure}[t]
    \centering
    \includegraphics[width=1\textwidth]{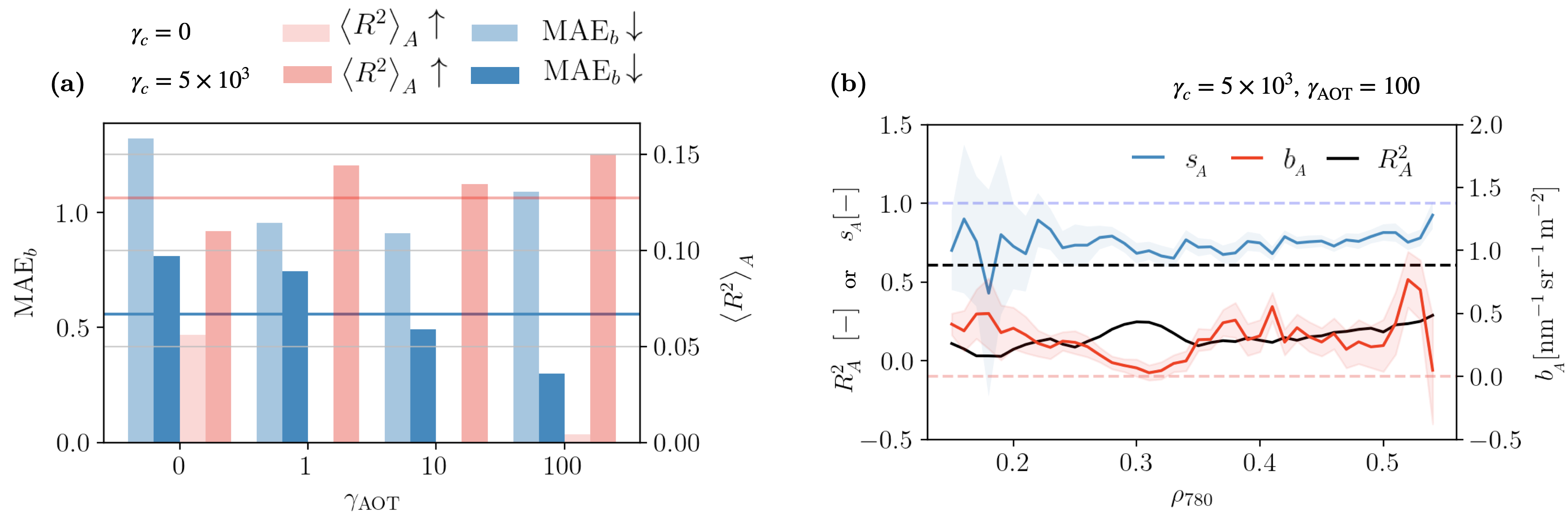}
        \caption{\textbf{(a): }Reflectance constrained metrics for $\gamma_\mathrm{AOT}$. Light colors indicate model runs with $\gamma_c = 0$, dark colors indicate runs with $\gamma_c = 5 \times 10^3$. Horizontal lines denote the reflectance constrained performance results of the model runs without ancillary data in \cref{fig:gamma_c_comparison} (d). \textbf{(b):} Reflectance constrained metrics for $\gamma_\mathrm{AOT} = 100$, $\gamma_c = 5\times 10^3$. Light blue and light red denote the uncertainty of the least-squares fit to gain $s$ and $b$. }
        \label{fig:gamma_aot_comparison_reflc}
\end{figure}

To assess the impact of including ancillary data in the SIF retrieval, we conduct a grid search over $\gamma_\mathrm{AOT}$ (while fixing $\gamma_\mathrm{H_2O} = 1$).
First, we establish the performance difference between using the proposed regularization scheme in \cref{eq:ell_m} and providing the data directly to the input and the emulator.
We denote this configuration by $\gamma_\mathrm{AOT} = 0$ in \cref{fig:gamma_aot}.
We find decreased $r^2$ and $\langle\Delta f_{740}\rangle$ performance in all HyPlant configurations (\cref{fig:gamma_aot}) compared  to cases $\gamma_\mathrm{AOT} > 0$.
Equally, decreased $r^2$ performance can be observed in the comparison with \mbox{OCO-3} SIF estimates. 
Secondly, we can observe that runs with high $\gamma_\mathrm{AOT}$ approximately reach the same SIF prediction performance as model runs without any ancillary data from \cref{sec:results_lC} (\cref{fig:gamma_aot} (a) and (c)). 
Finally, similarly to the results in \cref{sec:results_lC}, we can observe a performance increase in the HyPlant (2023) and \mbox{OCO-3} datasets when using $\gamma_c$. 
This is observable in $r^2$, $\langle \Delta f_{740}\rangle$,  $\mathrm{MAE}_b$ and $\left\langle R^2\right\rangle_{A}$ (\cref{fig:gamma_aot_comparison_reflc} (a)).
In particular, the previously observed large overestimation at high $\rho_{780}$ is reduced (\cref{fig:gamma_aot_comparison_reflc} (b)).

\section{Discussion}


\subsection{Decorrelating Impact of Consistency Regularization}
The introduction of $\ell_c$ proved beneficial to the overall SIF retrieval performance.
In a study of the performance of $\left\langle R^2\right\rangle_{A}$, we isolated the model performance independent of the empirical correlation between SIF and $\rho_{780}$.
We found improved performance in terms of $\left\langle R^2\right\rangle_{A}$ under the inclusion of $\ell_c$ (\cref{fig:gamma_c_comparison} (c) and (d)) indicating that the consistency regularization $\ell_c$ has the intended effect of decorrelating the target signal from confounding factors.
The weighting $\gamma_c$ has to be chosen carefully, however.
Increased  MAE$_b$ at large $\gamma_c$ may have been due (i) to imperfect sample generation in $\ell_c$ introducing a domain gap between the observations and the augmentations and (ii) a trade-off between reconstruction accuracy and minimization of $\ell_c$. 

\subsection{Inclusion of Ancillary Data Sources}
In order to reduce the retrieval problem's ill-posedness we have proposed the use of a regularization that implements the supervised learning of atmospheric emulator prediction variables with ancillary data sources as labels.
The regularization formulates a secondary downstream task in addition to the decomposition of the spectral observations into constituent variables.
Importantly, we could see improved SIF retrieval performance with this regularization in terms of a reduced \textit{reflectance dependent bias} MAE$_b$ when used with $\ell_c$.
As systematic integration of ancillary data is also planned for the FLEX mission by operating it in tandem with Sentinel-3 \cite{druschFLuorescenceEXplorerMission2017} 
the retrieval approach explored in this contribution could benefit similar SIF retrieval approaches on FLEX imagery.



\section{Conclusion}

In this contribution we have presented a deep learning architecture for SIF retrieval from DESIS imagery.
This work is the first to use hyperspectral DESIS data for SIF retrieval.
A unique data set of spatially and temporally closely matching HyPlant SIF estimates has allowed us to perform a detailed validation study of the methodology proposed in this work.
The good performance of our model with respect to these high-quality SIF estimates ($\left\langle \Delta f_{740} \right\rangle = 0.78 \, \,  \mathrm{mW\,  nm^{-1} \,  sr^{-1} \, m^{-2}}$, $r^2 = 0.6$) supports our finding that it is possible to derive SIF from DESIS products.
Further comparison with a data set of globally distributed \mbox{OCO-3} SIF estimates could establish the sensitivity of our SIF product in a wider variety of observational and ground conditions and may form the basis for an operational SIF product from DESIS data.

To achieve the good SIF prediction performance, we have extended a self-supervised simu\-lation-based deep learning approach \cite{buffatDeepLearningBased2023,buffatMultiLayerPerceptron2024}.
Several changes to the loss formulation were necessary to address the lower SR and SNR of DESIS imagery.
Most importantly, we have (i) introduced a perturbation based augmentation to improve signal decorrelation and (ii) tested the inclusion of ancillary data by formulating a secondary supervised downstream task.
We could show that both the perturbation based augmentation and the supervised downstream task formulations improved SIF retrieval performance when comparing both with HyPlant and \mbox{OCO-3} SIF products.
We furthermore could observe improved decorrelation of DESIS SIF from $\rho_{780}$ when making use of the augmentation during training.
Since this perturbation based regularization strategy is not restricted to remote sensing data it may be implemented in other simulation-based deep learning applications to decrease the influence of confounding factors.

\section*{Acknowledgements}
This work is part of the project ``FluoMap'' (Impulsfonds-Förderkenn\-zeich\-en ZT-I-PF-5-12) funded by the Helmholtz Initiative and Networking Fund, Helmholtz AI, Deutsches Zentrum für Luft- und Raumfahrt (DLR) and Forschungszentrum Jülich GmbH (FZJ). 
The authors gratefully acknowledge computing time on the supercomputer JURECA \cite{JURECA} at Forschungszentrum Jülich under grant no. fluomap-ct.

%
%

\end{document}